\documentclass{IEEEtran}
\usepackage[noadjust]{cite}
\ifCLASSINFOpdf
\else
\fi
\usepackage[T1]{fontenc} % optional enhanced font encoding
\usepackage[cmintegrals]{newtxmath}
\usepackage{bm}
\usepackage{array}
\usepackage{url}
\usepackage{cite}
\usepackage{amsmath,amsfonts}
\usepackage{algorithm}
\usepackage{algpseudocode}
\usepackage{graphicx}
\usepackage{textcomp}
\usepackage{xcolor}
\usepackage{balance}
\usepackage{wrapfig}
\usepackage{float}
\usepackage{subfig}
\usepackage{times}
\usepackage{caption}
\usepackage{makecell}
\usepackage{multirow}

\begin{document}

\title{Automated Defect Detection for Mass-Produced Electronic Components Based on YOLO Object Detection Models}

\author{Wei-Lung Mao, Chun-Chi Wang, Po-Heng Chou, \IEEEmembership{Member, IEEE}, and Yen-Ting Liu \vspace{-0.1in}
\thanks{Manuscript received September xx; revised xx; accepted xx. Date of publication xx; date of current version xx. This work was supported in part by Academia Sinica under Grant 235g Postdoctoral Scholar Program and in part by the National Science and Technology Council (NSTC) of Taiwan under Grant 113-2926-I-001-502-G. \emph{(Corresponding author: Po-Heng Chou and Chun-Chi Wang)}.}
\thanks{Wei-Lung Mao and Chun-Chi Wang are with the Department of Electrical Engineering and Graduate School of Engineering Science and Technology, National Yunlin University of Science and Technology (NYUST), Yunlin, 64002, Taiwan (e-mail:wlmao@yuntech.edu.tw, d11010202@gemail.yuntech.edu.tw).}
\thanks{Po-Heng Chou is the Research Center for Information Technology Innovation (CITI), Academia Sinica (AS), Taipei, 11529, Taiwan (e-mail: d00942015@ntu.edu.tw).}
\thanks{Yen-Ting Liu is with the Institute of Communications Engineering (ICE), National Sun Yat-sen University (NSYSU), Kaohsiung, 80424, Taiwan (e-mail: m113070007@student.nsysu.edu.tw).}}
%\thanks{Manuscript received XXX, XX, 2023; revised XXX, XX, 2023.}

%\markboth{IEEE Wireless Communications Letters,~Vol.~XX, No.~XX, XXX~2024}
{}
%{Shell \MakeLowercase{\textit{et al.}}: Bare Demo of IEEEtran.cls for Journals}

\maketitle
\begin{abstract}
 Since the defect detection of conventional industry components is time-consuming and labor-intensive, it leads to a significant burden on quality inspection personnel and makes it difficult to manage product quality. In this paper, we propose an automated defect detection system for the dual in-line package (DIP) that is widely used in industry, using digital camera optics and a deep learning (DL)-based model. The two most common defect categories of DIP are examined: (1) surface defects, and (2) pin-leg defects. However, the lack of defective component images leads to a challenge for detection tasks. To solve this problem, the ConSinGAN is used to generate a suitable-sized dataset for training and testing. Four varieties of the YOLO model are investigated (v3, v4, v7, and v9), both in isolation and with the ConSinGAN augmentation. The proposed YOLOv7 with ConSinGAN is superior to the other YOLO versions in accuracy of 95.50\%, detection time of 285 ms, and is far superior to threshold-based approaches. In addition, the supervisory control and data acquisition (SCADA) system is developed, and the associated sensor architecture is described. The proposed automated defect detection can be easily established with numerous types of defects or insufficient defect data.
\end{abstract}

\begin{IEEEkeywords}
Automated defect detection, automated optical inspection (AOI), camera sensor, ConSinGAN, deep learning (DL), dual in-line package (DIP) switch, you only look once (YOLO). 
\end{IEEEkeywords}

\IEEEpeerreviewmaketitle

\section{Introduction}
\label{sec:introduction}
\IEEEPARstart{G}{lobal} market trend~\cite{re1_Dip_switches_Market} shows that the global market size of the dual in-line package (DIP)~\cite{patent:5010445} switch achieved USD 400.1 million in 2021.
Based on a compound annual growth rate of 3.3\% during the forecast period, it is predicted to reach USD 554.35 million by 2031. 
It exhibits a steady annual growth with approximately 10 million units produced daily.
With this huge amount of the manufacturing process, it is an essential issue to manage product quality and yield rates.
However, conventional quality management depends on human visual product inspections.
Due to the time-consuming and labor-intensive nature of the task, it leads to a significant burden on quality inspection personnel.
Therefore, the current trend of Industry 4.0 is to streamline manufacturing industries, converting from labor-intensive work toward deep learning (DL)-based automated production.

%{\color{red} Several basic components such as resistors and inductors, DIP switches are also widely used in various electronic products and printed circuit boards (PCBs). In particular, the DIP switch, commonly used on a printed circuit board, is adopted as the testing component. However, the DIP inspection process is inspected manually, which is complicated. With daily production demands reaching millions and output increasing year by year, automated inspection is inevitable.

Over the past decade, DL~\cite{re_add_11,re_add_12,re_add_13,re_add_14} has been widely applied to various fields, including image recognition, natural language processing, autonomous driving, and object classification problems. Further, DL has also been used for medical radiography image detection and classification~\cite{re2,re3,re4,re5}, A. Jo and W. Lee,~\cite{re6} proposed a DL-based model for material discrimination and quantitation. S. M. J. Jalali~\emph{et al}.,~\cite{re7} proposed the deep reinforcement learning-based ensemble algorithm that integrates optimized deep learning models to minimize wind power forecasting errors for two wind power datasets. In~\cite{re10}, a DL-based model was proposed by E. Jeong~\emph{et al}. for heterogeneous inference parallelization.

Several works~\cite{re12,re13,re14} use DL to reduce labor resource consumption and promote production line automation. In~\cite{re12}, a dual-weighted analysis algorithm was proposed by  K. Qiu \emph{et al}. for metal defect detection. Y. Chen~\emph{et al}.~\cite{re13} focused on improving the metal punching position using fiber Bragg grating (FBG) sensors and identifying the impact source location through a hyperbolic localization model. In~\cite{re14}, A. Albanese~\emph{et al}. developed a system using a micro-camera control unit (MCU) for automatic DL imaging and detection. Several studies~\cite{re16,re17,re19,re20} use DL and threshold-based methods to detect defects. L. Xiao \emph{et al}.~\cite{re16} utilized wavelet transformations to process image features for surface defect detection. Y. Zhang \emph{et al}.~\cite{re17} formulated defect detection as a classified problem, which could then be resolved via support vector machines (SVM) and random forests (RF). S. Tian \emph{et al}.~\cite{re19} proposed a complementary adversarial network-driven surface defect detection (CASDD) framework to identify different types of texture defects automatically and accurately. Y. Peng \emph{et al}.~\cite{re20} proposed a method to detect surface defects on electric distribution cabinets using SVM and Gaussian distribution modeling threshold techniques. A. Dairi \emph{et al}.~\cite{re_add_9} proposed a novel stereo-vision method that merges a deep encoder with $k$ nearest neighbor (KNN) for anomaly detection to detect obstacles in a road environment. In ~\cite{re_add_10}, A. Dairi \emph{et al}. proposed a deep encoder with a one-class support vector machine (OCSVM) for obstacle detection systems. Table~\ref{add_tb1_Comparison_methods} summarizes the applications and objects of the methods above. %Our objective is to use image detection to identify defects in DIP-type components.

\begin{table}[htbp]
\centering
\caption{Comparison of defect detection methods}
\scalebox{0.83}{
      \begin{tabular}{|c|c|c|c|}
      \hline
        \textbf{Methods} & \textbf{\makecell{Model / Algorithm}} &\textbf{Detected objective}&\textbf{Goal}\\
        \hline
        \textbf{\multirow{6}{*}{DL}}  & \multirow{3}{*}{CNN} & Test components~\cite{re14} & Surface defect \\\cline{3-4}
        &  & Test components~\cite{re13} & Object positioning\\\cline{3-4}
        &  & Steel~\cite{re19} &Surface defect \\\cline{2-4}
        &IPCNN & Car~\cite{re16} &Oil leak detection \\\cline{2-4}
        &KNN & \multirow{2}{*}{Obstacle~\cite{re_add_9,re_add_10}} &\multirow{2}{*}{Obstacle detection} \\\cline{2-2}
        &OCSVM & & \\\hline 
        \textbf{\multirow{3}{*}{Threshold}}  &\makecell{Wavelet multiscale\\analysis}& Tire~\cite{re17} & Surface defect\\\cline{2-4}
        &\makecell{Gaussian distribution\\modeling,\\SVM}  & \makecell{Electric distribution\\cabinet~\cite{re20}}&Surface defect  \\\cline{2-4}
        &\makecell{Dual-weighted\\analysis algorithm}  & Metal parts~\cite{re12} &Surface defect  \\\hline    
        \end{tabular}}
     \label{add_tb1_Comparison_methods}
\end{table}
However, these studies overlooked several key aspects: 
1) Real-time detection: Most studies focus on performance in terms of accuracy, and neglect the detection speed performance in the practical application.
2) Computational limitations: It is crucial to lighten the DL model to fit hardware constraints, rather than increasing its size indefinitely for computational capability.
3) Insufficient dataset: The lack of datasets leads to a challenge for detection tasks. 

Therefore, we design the lightweight DL model to fit hardware limitations and real-time detection capability. 
From the review of the image detection models~\cite{re_add_1,re_add_2,re_add_3,re18}, we selected anchor-based one-stage models, specifically the You Only Look Once (YOLO)~\cite{re25_YOLOv3,re26_YOLOv4,re27_YOLOv7,re_add_5_YOLOv5, re_add_6_YOLOv6,re_add_7_YOLOv8,re_add_8_YOLOv9}.
 
YOLO is a neural network that has shown superior performance in terms of both real-time operation and accuracy. YOLO uses the feature pyramid network (FPN) framework enhancement for iterative updates to achieve better results until convergence. YOLO has demonstrated commendable performance in various high-demand real-time scenarios such as public face recognition, mask detection, self-driving car systems, automated manufacturing systems, surface defect detection, etc.
Several studies~\cite{re28,re29,re30,re31,re32,re33,re34,re35,re36,re37,re38,re39,re40} suggested using YOLO for the dynamic detection of metal surface defects~\cite{re28,re29,re31,re32}. The authors of~\cite{re30} applied YOLO to real-time detection for traffic signs. The authors of~\cite{re37} performed a visual detection of autonomous robots by YOLO. Furthermore, improved YOLO models were used to detect surface defects in solar cells in~\cite{re34} and~\cite{re38}. YOLO has also been used for real-time detection of ships~\cite{re33,re35} and vehicles~\cite{re36,re39,re40}.

To the best of our knowledge, this work is the first study to investigate DL-based automated DIP image detection.
A quality-control defect detection method is proposed for each side of this hexagonal component.
The proposed system is not a stand-alone system but is integrated with automated manufacturing production line processes.
Therefore, the detection time of the automated production output from the preceding workstation must be considered to ensure that detection is completed within a specified time frame.

On the other hand, the primary challenge of automated image processing is the scarcity of defective product samples.
To enhance the limited amount of samples, a generative adversarial network (GAN) model~\cite{re21} called ConSinGAN~\cite{re22} is used to increase the input image quantity for data argumentation.
The ConSinGAN can develop a model based on a single image and effectively simulate the defect characteristics in the generated image.
It greatly facilitates the training process of detection models. 
The other GANs, such as deep convolutional GAN (DCGAN)~\cite{re23_DCGAN} or Wasserstein GAN (WGAN)~\cite{re24_WGAN}, require more images for training.
The main contributions of this paper are described as follows:
\begin{itemize}
    \item The proposed defective DIP detection system is designed based on the YOLO model to improve the quality of the inspection.
    \item To enhance the performance of the YOLO model, ConSinGAN is used to generate the amount of DIP image data for the training phase.
    \item To validate the effectiveness, we develop a practical production line and SCADA interface to reflect the performance in practical applications.
    \item To compare the different YOLO versions, we evaluate the performances in terms of accuracy and detection time.
    \item Compared with threshold-based detection, the proposed YOLO models with ConSinGAN reveal a reduction in production detection time by 909-948 ms, where the YOLOv7 with ConSinGAN achieves an accuracy of 95.50\%.
    \item Unlike previous studies that only focused on detection models, we implement the proposed detection in a practical automated mechanism, while considering the detection time for performance measurement.
\end{itemize}

The remainder of this paper is organized as follows. 
Sec.~\ref{sec:Automated system development} presents the architecture of the automated defect detection system, the types of DIP defects, and threshold-based image detection for baseline comparison. Sec.~\ref{sec:Detection system} details the data pre-processing, YOLO models, and performance metrics. Sec.~\ref{sec:Experimental results} compares the simulation results among the threshold-based detection, DL-based YOLOv3, v4, v7, and v9 models with/without ConSinGAN for DIP defect detection. Sec.~\ref{sec:Conclusion} summarizes the conclusions and the future works.

\section{Automated Defect Detection System}
\label{sec:Automated system development}
\begin{figure}[h]
    \centerline{\includegraphics[width=0.6\columnwidth]{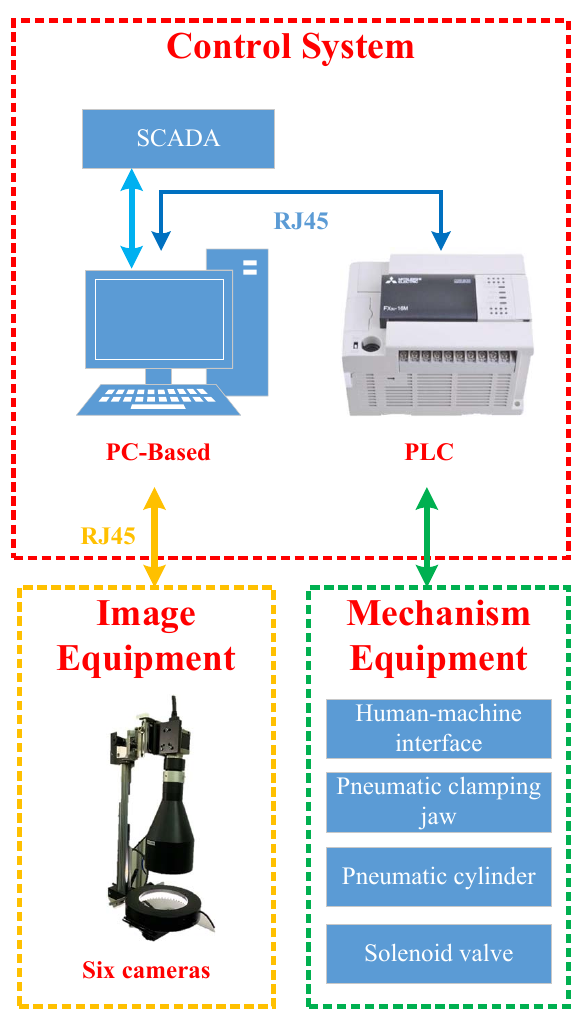}}
    \caption{System architecture diagram.}
    \label{fig:1}
\end{figure}

The system architecture is shown in Fig.~\ref{fig:1} and includes three parts: 1) Control system: Comprised of a personal computer (PC) and the programmable logic controller (PLC) devices. The PC is used to interact with the imaging equipment and establish the SCADA, which allows for integration with the DL model and data analysis. The PLCs are used to connect the mechanical equipment, control various actions, issue action completion signals, and provide a secondary confirmation count. 2) Imaging equipment: industrial cameras, centrifugal lenses, and various light source equipment. The imaging equipment interacts and feeds data to the PC via an Ethernet cable (RJ45). 3) Mechanical equipment: The human-machine interface of the PLC includes pneumatic clamps, electromagnetic push rods, and solenoid valves.

\subsection{Automated Mechanism}
\label{secsub:Automated mechanism}
\begin{figure}[h]
\centerline{\includegraphics[width=0.7\columnwidth]{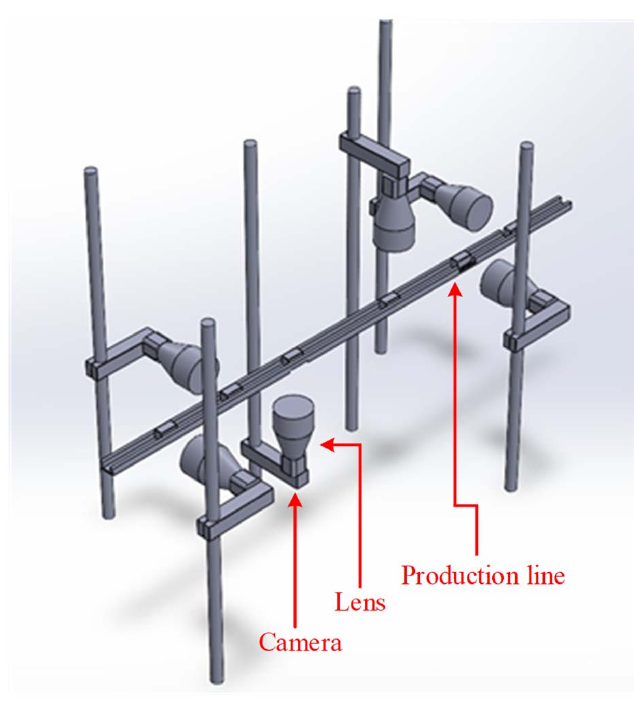}}
\caption{3D simulation schematic.}
\label{fig:2}
\end{figure}

The proposed system is designed to inspect all six sides of the workpiece. Each side is equipped with multiple Automated Optical Inspections (AOI) at different locations to minimize interference from various light sources on different surfaces. A 3D schematic diagram of the proposed system architecture is shown in Figure~\ref{fig:2}. Fig.~\ref{fig:2} shows the position and orientation of the AOI and the design of the experimental fixture. It establishes the positioning of lenses and light sources at different angles to inspect the long and short sides of a rectangular workpiece. The detected DIP is placed inversely at the production line. Thus, the DIP top-side detection uses the camera to capture images from bottom to top, and the DIP bottom-side detection captures images from top to bottom. The back- and front- sides of DIP are shot using two sets of cameras from the left and right. Side-to-side imaging is facilitated by rotating the device, allowing cameras on both sides to capture images. The longer sides are the up, down, back, and front parts of the detected DIP. The shorter sides are the left and right of the detected DIP. To examine the longer and shorter surfaces of rectangular DIP, the experiment required two sets of cameras and lenses, each with a specific depth of field. Long-side cameras use a low depth of field, and short-side cameras use a high depth of field. Tables ~\ref{tb1:low depth of field} and ~\ref{tb2:High} list the specific parameters of the camera lens set.

\begin{table}[h]
\centering
\caption{Low depth of field camera and lens specifications}
\scalebox{0.8}{
        \setlength{\tabcolsep}{1mm}
        \begin{tabular}{|c|c|c|c|}\hline
        \multicolumn{2}{|c|}{BASLER-acA4096-30 µm (Camera)} &  \multicolumn{2}{c|}{SPO-200I-4M (Len)}\\
        \hline
        \textbf{Specification}  & \textbf{Parameter} & \textbf{Specification}  & \textbf{Parameter}
        \\\hline
        \textbf{Pixel size}  & 3.45 µm & \textbf{Depth of field}  & 4.5 
        \\\hline
        \textbf{Image resolution}  & 4096$\times$2168 & \textbf{Working distance}  & 200   
        \\\hline
        \textbf{Magnification}  & 0.231 & \textbf{Field of view
(FOV)}  & 61.0$\times$33.4        
        \\\hline
    \end{tabular}}
\label{tb1:low depth of field}
\end{table}

\begin{table}[h]
\centering
\caption{High depth of field camera and lens specifications}
\scalebox{0.8}{
      \setlength{\tabcolsep}{1mm}
      \begin{tabular}{|c|c|c|c|}\hline
        \multicolumn{2}{|c|}{BASLER-acA3800-14 µm (Camera)} &  \multicolumn{2}{c|}{OPTO-TC23048(Len)}\\
        \hline
        \textbf{Specification}  & \textbf{Parameter} & \textbf{Specification}  & \textbf{Parameter}
        \\\hline
        \textbf{Pixel size}  & 1.67 µm & \textbf{Depth of field}  & 20.0 mm
        \\\hline
        \textbf{Image resolution}  & 3860$\times$2748 & \textbf{Working distance}  & 132.9 mm  
        \\\hline
        \textbf{Magnification}  & 0.184 & \textbf{Field of view
(FOV)}  & 34.9$\times$24.9 mm        
        \\\hline
    \end{tabular}}
\label{tb2:High}
\end{table}

The size of each workpiece is approximately 21.7 mm $\ pm$0.5 mm on the long side and 9.4 mm $\ pm$0.4 mm on the short side. The resolving power of the image equipment is calculated by
\begin{equation}
        RP = \frac{PS}{MF},
        \label{eq1} 
\end{equation}
where $RP$ is the resolving power, $PS$ is the pixel size, and $MF$ is the magnification. The field of view (FOV) is calculated by
\begin{equation}
        {\rm FOV} = RP \times IR, 
        \label{eq2}
\end{equation}
where $IR$ is the image resolution.

\subsection{Defective Types of DIPs}
\label{subsec:work defec}
\begin{figure}[h]
\centerline{\includegraphics[width=\columnwidth]{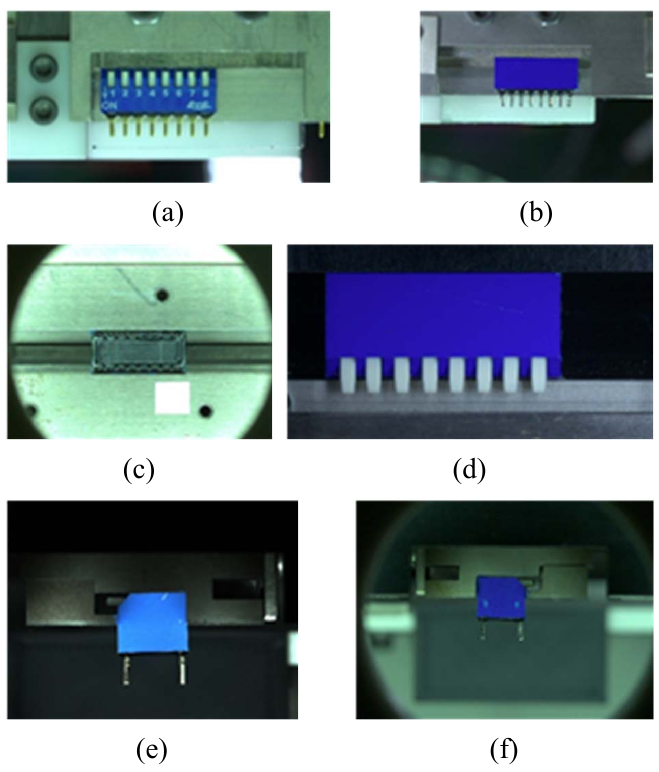}}
\caption{Different sides of Normal DIP, (a) front, (b) back, (c) bottom, (d) top, (e) right, and (f) left sides.}
\label{fig:3}
\end{figure}

\begin{figure}[h]
\centerline{\includegraphics[width=\columnwidth]{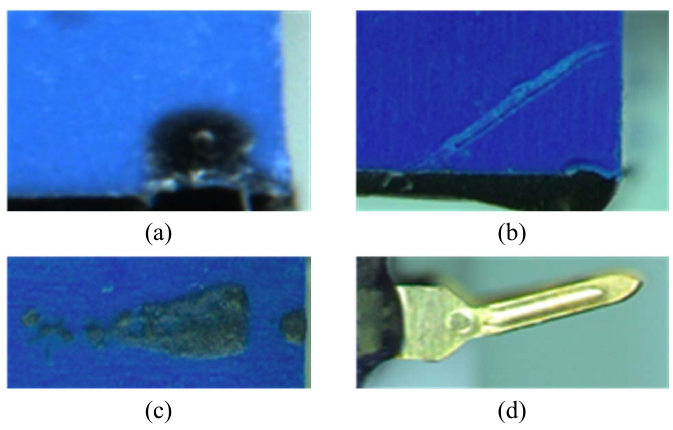}}
\caption{Four types of defective DIP, (a) surface glue overflow, (b) surface scratches, (c) surface dirt, and (d) bent pins.}
\label{fig:4}
\end{figure}

A normal DIP switch is shown in Fig.~\ref{fig:3}. For defective DIPs, there are two categories: surface and pin-leg. The surface defects include three types: surface overflow, surface scratches, and surface contamination. In addition, misaligned pin-legs are also a type of defect. The above four defective DIPs are shown in Fig.~\ref{fig:4}.

The sizes of the defect feature are as follows: Fig.~\ref{fig:4}(a) is approximately 100 to 300 pixels, Fig.~\ref{fig:4}(b) is approximately 100 to 500 pixels, Fig.~\ref{fig:4}(c) is approximately 100 to 300 pixels, and Fig.~\ref{fig:4}(d) is 220 to 250 pixels per pin. To enhance the contrast degree of the feature, the images in Fig.~\ref{fig:4}(a),~\ref{fig:4}(b), ~\ref{fig:4}(c), and ~\ref{fig:4}(d) can use the red green blue (RGB) channel subtraction, such as red (R) minus green (G) and G minus blue (B) to enhance the features of the image.

\subsection{Threshold-Based Image Detection}
\label{subsec:Threshold-based image detection}

For a baseline comparison, we introduce a threshold-based image detection. First, it performs binary classification of the original image using a preset threshold to obtain the featured image. The feature extraction is determined by
\begin{equation}
        b(x,y)= \begin{cases}
        255,\mathrm{MinGray} \le g(x,y)\le \mathrm{MaxGray}, \\
        0, \ \rm otherwise,
        \end{cases}
        \label{eq3}
\end{equation}
where $b(x,y)$ represents the pixel of the featured image based on the binary decisions, $g(x,y)$ is the pixel of the original image, $\rm MinGray$ is the minimum threshold, and $\rm MaxGray$ is the maximum threshold.

\begin{figure}[h]
\centerline{\includegraphics[width=0.7\columnwidth]{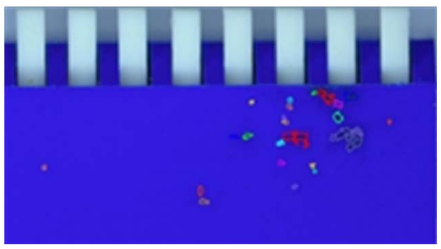}}
\caption{The defective features from threshold-based detection.}
\label{fig:5}
\end{figure}

Since the light intensity corresponding to each camera varies with the six sides of the DIP surface, the threshold values $\rm MinGray$ and $\rm MaxGray$ are adjustable.
Whether each side of the DIP surface is defeated or not is determined by
\begin{equation}
        DR(z)= \begin{cases}
       \mathrm{Defective \; DIP} ,\ TB\le o(z)\\
        \mathrm{Normal \; DIP}, \ TB>o(z)>0,
        \end{cases}
        \label{eq4}
\end{equation}
where $DR(z)$ represents the threshold-based detected result, $TB$ represents the preset threshold, $z$ denotes the featured image, and $o(\cdot)$ is the output image of feature extraction.
The features of defective DIP are shown in Fig.~\ref{fig:5}. 

\vspace{-0.1in}
\section{DL-based YOLO Model and ConSinGAN}
\label{sec:Detection system}

\begin{figure}[h]
\centerline{\includegraphics[width=0.5\columnwidth]{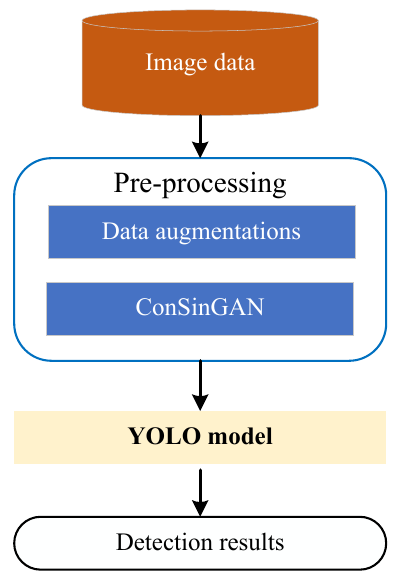}}
\caption{The flow chart of the proposed defective DIP detection.}
\label{fig:6}
\end{figure}

Fig.~\ref{fig:6} shows the workflow of the proposed detection.
To select the outstanding YOLO version, we compare different YOLO versions in terms of accuracy, floating-point operations (FLOPs), and the number of parameters.
To enhance the accuracy of the YOLO models, we augment the data set by using ConSinGAN~\cite{re22} during the data pre-processing.
The ConSinGAN modifies the original images using rotation, translation, flipping, scaling, Gaussian noise, etc, for data augmentation.

\subsection{Data Pre-Processing}
\label{subsec:Pre-processing}

\subsubsection{Data Augmentation}

\begin{figure}[h]
\centerline{\includegraphics[width=0.5\columnwidth]{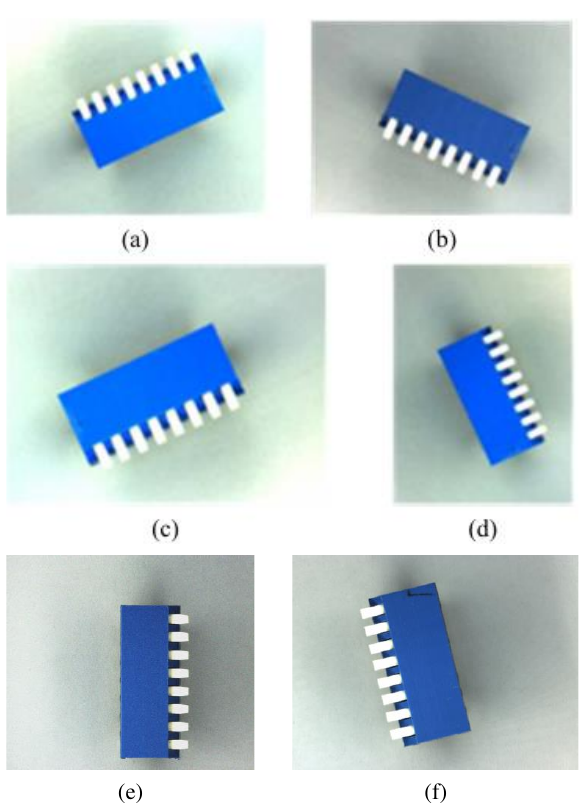}}
\caption{Image augmentation. (a) brightness adjustment, (b) mirroring, (c) median filtering, (d) flipping, (e) noise, and (f) Gaussian blur.}
\label{fig:7}
\end{figure}

To augment the defective DIP image dataset, various techniques are used: flipping, mirroring, brightness adjustment, median filtering, noise, and Gaussian blur, as depicted in Fig.~\ref{fig:7}. Data augmentation is crucial to improve the trained model and enhance the discrimination.

\subsubsection{ConSinGAN}
\begin{figure}[h]
\centerline{\includegraphics[width=1\columnwidth]{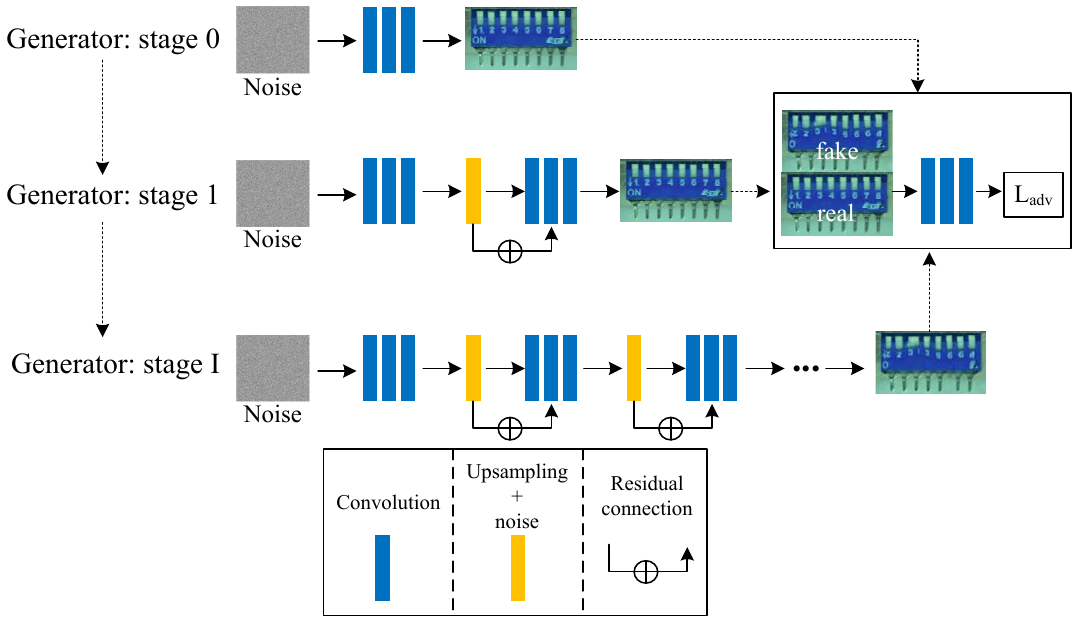}}
\caption{The concept of the ConSinGAN.}
\label{fig:ConSinGAN}
\end{figure}

The predecessor of ConSinGAN, SinGAN~\cite{re41}, introduces a GAN that is trained on a single image for tasks such as unconditional image generation and harmonization. The Concept of the ConSinGAN model is shown in Fig.~\ref{fig:ConSinGAN}.

ConSinGAN is trained using a multi-stage and multi-resolution approach with the lowest resolution (e.g., 25 × 25 pixels) in the first stage.
The next stage increases the layers of the neural network and image resolution.
At each stage, the layers of the neural network in all the previous stages are frozen, and only the current added layers of the neural network are trained. 
Different from the traditional GAN, it uses a multi-generator, and the number of stages increases the number of generators.
The traditional GAN propagates feature maps from one generator stage to the next, which leads to a negative effect on learning performance.
The training of ConSinGAN is limited to one stage at a specific time to prohibit interactions between different stages.
If all stages are end-to-end training, it causes overfitting in the single-image case, and the network collapses.
It means that the relationship between the receptive field and the size of the generated image decreases as the number of stages increases. 
At higher resolutions, the discriminator focuses on the texture of the image.
At lower resolutions, the discriminator concerns the global layout of the image.
At a given stage $i$, ConSinGAN optimizes the sum of an adversarial and a reconstruction loss, and initializes the discriminator with the weights of the previous stage $i-1$ at all stages as follows:
\begin{align}
        \min_{G_{i}} \max_{D_{i}} \mathcal{L}_{adv}(G_{i},D_{i})=\alpha\mathcal L_{rec}(G_{i}),
        \label{eq5}
\end{align}
where $\mathcal{L}_{adv}(G_{i},D_{i})$ is the adversarial loss that is the same with WGAN~\cite{re24_WGAN}, the reconstruction loss $L_{rec}(G_{i})$ is used to improve the training stability ($\alpha$ is set to 10 in our experiments) as follows:
\begin{equation}
        \mathcal{L}_{rec}(G_{i})=||G_{i}(s_{0})-s_{i}||_{2}^{2}.
        \label{eq6}
\end{equation}
At the given resolution of stage $i$, we input $(s_{0})$ into the generator $G_{i}(s_{0})$. The input $(s_{0})$ is a downsampling version of the original image $(s_{i})$.

\subsection{YOLO Object Detection Model Series}
\label{subsec: YOLO learning series}
YOLO is adopted as a DL-based object detection algorithm. Compared with traditional object detection algorithms, the YOLO model can quickly perform object detection and classification within a single neural network. The YOLO architecture is a convolutional neural network (CNN) that includes multiple convolutional and fully connected layers, segments an image into S × S grids, and then detects multiple objects in each grid. In addition, the selection criteria of the different YOLO versions for DIP testing were based on their floating-point operations (FLOPs) and mAP. To compare different YOLO versions, we list the performance in terms of FLOPs, and of YOLO v3, v4, v5, v6, v7, v8, and v9, as shown in Table~\ref{tb:Comparison of the YOLO series}. YOLOv7~\cite{re27_YOLOv7} is chosen as the proposed DIP testing model. In our simulation, YOLOv3~\cite{re25_YOLOv3}, YOLOv4 ~\cite{re26_YOLOv4}, and the current newest YOLOv9~\cite{re_add_8_YOLOv9} are considered for comparison.
As shown in Table~\ref{tb:Yolo model results}, the YOLOv7 slightly outperforms YOLOv9 in terms of accuracy but is slightly slower. YOLOv5~\cite{re_add_5_YOLOv5} and YOLOv6~\cite{re_add_6_YOLOv6} have similar performance to YOLOv7 in terms of FLOPs, but their mAP is less than that of YOLOv7. YOLOv8 has higher computation complexity than YOLOv7 in terms of FLOP by 63\%; however, the mAP improved by only 0.1\%. Therefore, YOLOv5, v6, and v8 are excluded from our main simulations.
\begin{table}[h]
\centering
\caption{Comparison of the YOLO series}
\scalebox{0.8}{
      \begin{tabular}{|c|c|c|c|}
      \hline
       \textbf{Model} & \textbf{\#Param (M)} & \textbf{FLOPs} &\textbf{mAP0.5} \\\hline
       \textbf{YOLOv3~\cite{re25_YOLOv3}} & 62 & 1457B &57.9\%\\\hline
       \textbf{YOLOv4~\cite{re26_YOLOv4}} & 53 & 109B &65.7\%\\\hline
       \textbf{YOLOv5-L~\cite{re_add_5_YOLOv5}} & 46.5 & 109.1G &67.3\%\\\hline
       \textbf{YOLOv6-M~\cite{re_add_6_YOLOv6}} & 34.9 & \textbf{85.8G} &66.4\% \\\hline
       \textbf{YOLOv7~\cite{re27_YOLOv7}} & 36.9 & 104.7G &69.7\% \\\hline
       \textbf{YOLOv8-L~\cite{re_add_7_YOLOv8}} & 43.7 & 165.2G &69.8\%\\\hline
       \textbf{YOLOv9-C~\cite{re_add_8_YOLOv9}} & \textbf{25.3} &102.1G &\textbf{70.2\%}\\\hline
       \end{tabular}}
\label{tb:Comparison of the YOLO series}
\end{table}

To further compare the YOLOv3, v4, v7, and v9 models, we introduce the upgraded parts of these models in the following: 
\subsubsection{YOLOv3~\cite{re25_YOLOv3}}
The most significant improvement of YOLOv3 is adopting the multiscale feature extraction, also known as a feature pyramid network (FPN). 
In addition, YOLOv3 uses binary cross-entropy as the loss function, instead of the mean square error of YOLOv2~\cite{re_add_16_YOLOv2}.
YOLOv3 uses Darknet-53 to replace darknet-19 of YOLOv2 and increase depth and accuracy, but it sacrifices the detection speed. 

\subsubsection{YOLOv4~\cite{re26_YOLOv4}}
First, YOLOv4 uses CSPDarknet53 to establish the primary backbone neural network, making it a lightweight model while maintaining high accuracy.
Second, YOLOv4 adopts spatial pyramid pooling (SPP)~\cite{re43} and path aggregation network~\cite{re44} in the neck section to combine feature maps at different scales and detect different sizes of objects.
Finally, the YOLOv4 employs complete intersection over union (CIoU)-loss to evaluate the distance between the targeted and the predicted boxes and improve the accuracy.

\subsubsection{YOLOv7~\cite{re27_YOLOv7}}
\begin{figure*}[h]
    \centerline{\includegraphics[width=2.1\columnwidth]{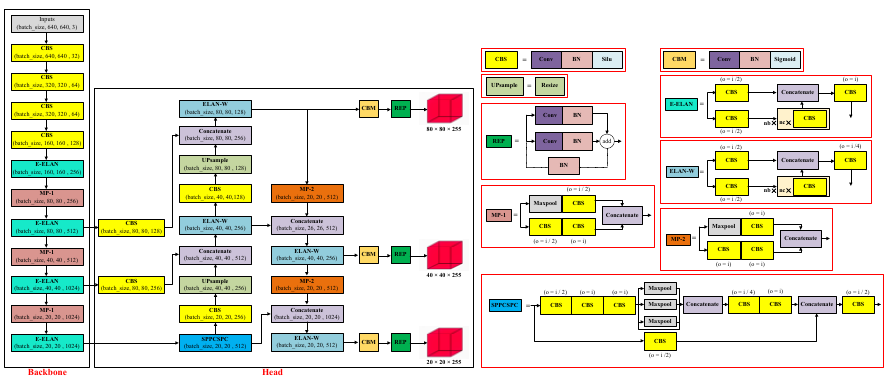}}
    \caption{The YOLOv7 network architecture and block description.}
    \label{Fig:YOLOv7}
\end{figure*}

YOLOv7~\cite{re27_YOLOv7} and YOLOv4~\cite{re26_YOLOv4} were developed by H.-Y. M. Liao,~\emph{et al}.
The backbone of YOLOv7 is redesigned based on v4.
The backbone is used to extract the features from the input images.
Unlike YOLOv5 and v6, YOLOv7 is not pre-trained on ImageNet. 
YOLOv7 has several new architectures with improved performances in terms of accuracy and detection speed.
The architecture of YOLOv7 is shown in Fig.~\ref{Fig:YOLOv7}

\begin{figure}[h]
    \centerline{\includegraphics[width=0.7\columnwidth]{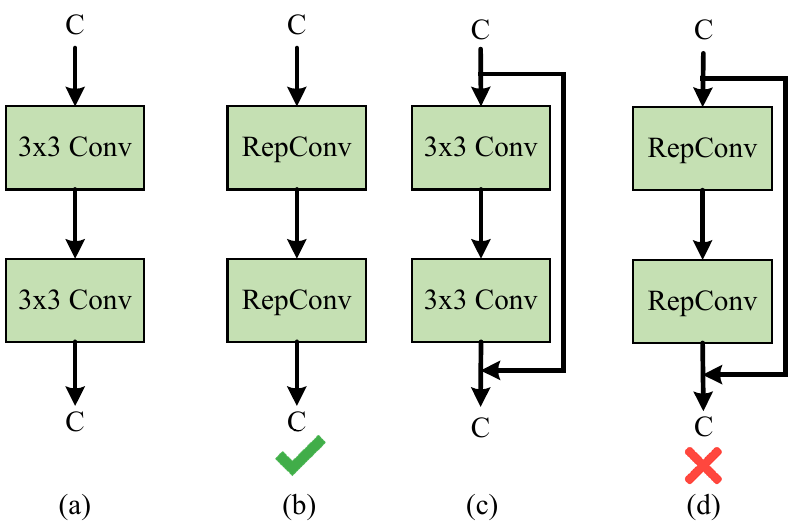}}
    \caption{The proposed RepConvN using (a) PlainNet, (b) RepPlainNet, (c) ResNet, and (d) RepResNet.}
    \label{fig:8}
\end{figure}

In Fig.~\ref{Fig:YOLOv7}, YOLOv7 uses an extended efficient layer aggregation network (E-ELAN), which develops the original ELAN architecture through expanding, shuffling, and merging.
These approaches allow the network to learn more diverse features without disrupting the original gradient paths.
The head part of YOLOv7, which is used for object detection and localization, continues to utilize the head network structure of YOLOR~\cite{re_add_15_YOLOR}.
The convolution batch normalization silu (CBS) block includes the convolutional layer, batch normalization (BN), and the Mish activation function~\cite{mish_bmvc}.
The spatial pyramid pooling cross-stage partial (SPPCSPC) is the first module in the head region, and it receives input from E-ELAN in the backbone. The SPPCSPC is composed of a cross-stage partial network (CSPNet) with an SPP block instead of a dense block as in YOLOv5~\cite{re_add_5_YOLOv5}.
The CBM block consists of the convolutional layer, batch normalization, and the sigmoid-weighted linear unit (silu) activation function. The reparameterization (REP) block represents the RepVGG.
The difference between max pooling (MP)-1 and MP-2 is the output size within the block.
The difference between E-ELAN and ELAN-W is the output size within the block.
The main improvements of YOLOv7 are as follows:

\begin{itemize}
\item In re-parameterized refocusing convolution (RepConv), the identity connection (id) disrupts the residual in the residual neural network (ResNet) and the concatenation in dense convolutional network (DenseNet). However, RepConv needs to retain features that provide greater gradient diversity for different feature maps. A novel weight parameterization approach is proposed in~\cite{re45} for RepConv to replace id. 
In Fig.~\ref{fig:8}, the planned reparametrized convolution designed for ResNet is called RepConvN (RepConv without identity connection (id))~\cite{re27_YOLOv7}.
Since PlainNet in Fig.~\ref{fig:8}(a) inherently lacks residual or concatenation characteristics, RepConv in Fig.~\ref{fig:8}(b) can be applied directly. 
Since ResNet in Fig.~\ref{fig:8}(c) inherently has residual characteristics, RepConv in Fig.~\ref{fig:8}(d) leads to a decrease in accuracy.
Based on the above reasons, YOLOv7 uses RepConvN to design a network architecture.
\end{itemize}

\begin{figure}[h]
    \centerline{\includegraphics[width=0.7\columnwidth]{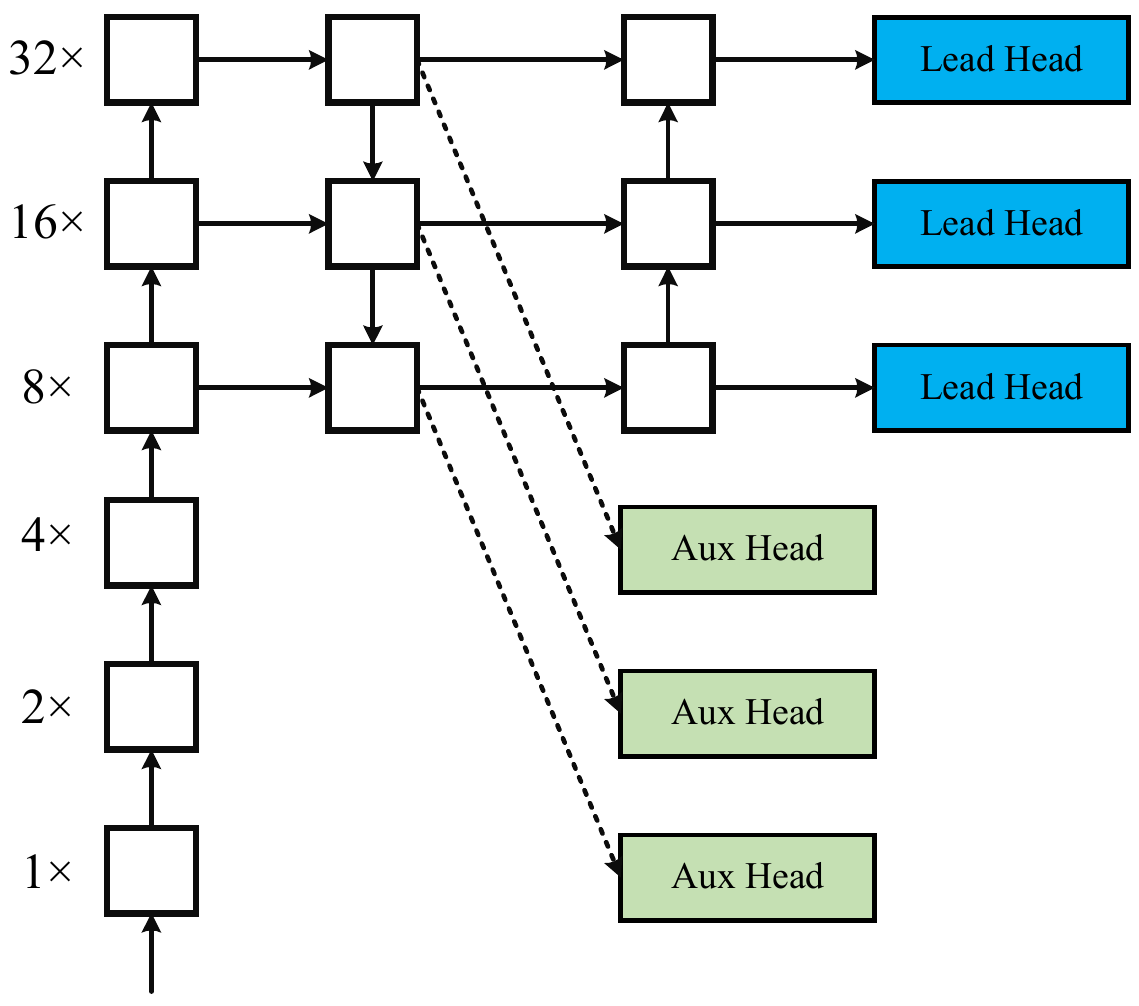}}
    \caption{YOLO model with auxiliary head.}
    \label{fig:9}
    \end{figure}

\begin{itemize}
\item Supervised learning is commonly used to train deep networks, where the weights of shallower layers are trained using an auxiliary mechanism to improve the accuracy of YOLOv7.
In addition, YOLOv7 introduces a novel dynamic label assignment strategy that employs a coarse-to-fine guide to enhance feature learning.
The final output head is called the lead head, and the head used for auxiliary training is called the auxiliary head.
Fig.~\ref{fig:9} depicts an object detector equipped with an auxiliary head.
\end{itemize}

\begin{figure}[h]
    \subfloat[\label{subfig:10 (a)}]{%
    \includegraphics[width=0.225\textwidth]{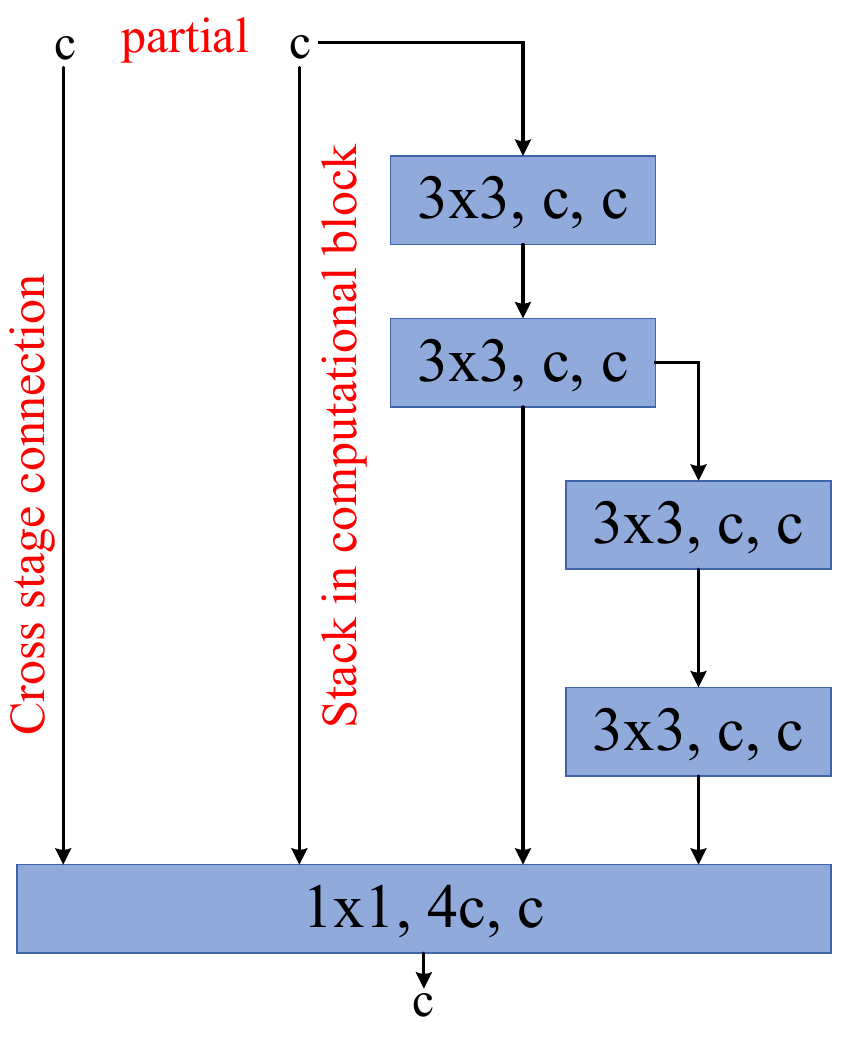}
    }
   \hfill
    \subfloat[\label{subfig:10 (b)}]{%
    \includegraphics[width=0.25\textwidth]{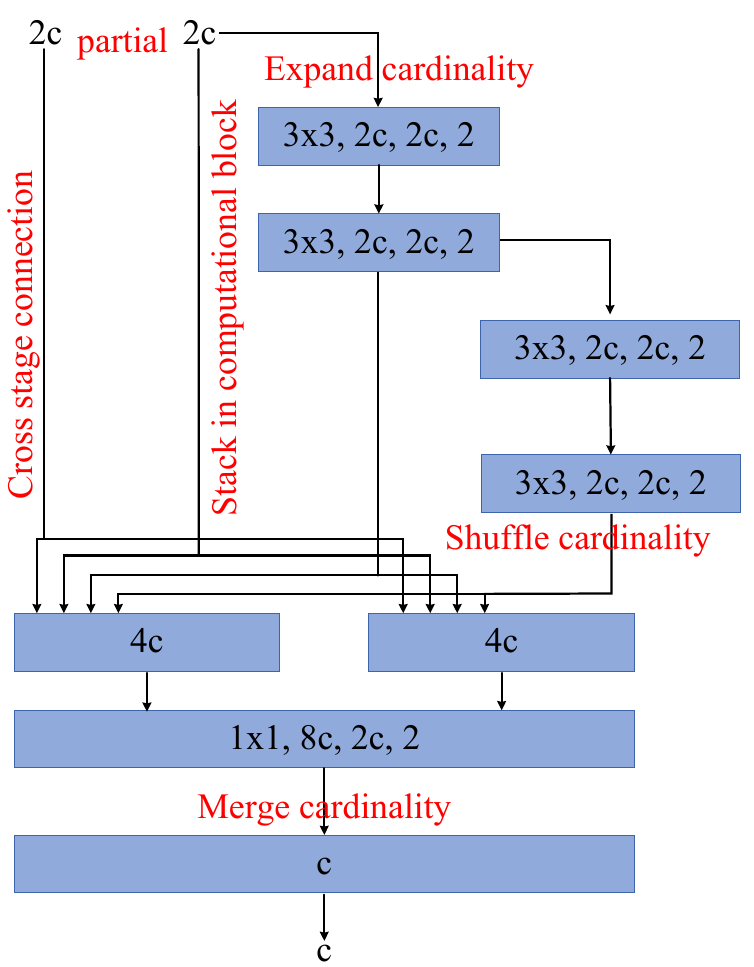}
    }
    \caption{ELAN optimization schematic for (a) ELAN, and (b) E-ELAN.}
    \label{fig10}
\end{figure}

\begin{itemize}
\item By optimizing ELAN, as shown in Fig.~\ref{fig10}(a), the shortest and longest gradient paths are controlled to allow deeper layers to learn and converge more effectively.
However, stacking more CNN layers in ELAN  may disrupt stable learning.
Therefore, YOLOv7 proposes extended ELAN (E-ELAN) in Fig.~\ref{fig10}(b) to maintain gradient paths while continuously enhancing the learning performance using expanding, shuffling, and merging of cardinalities.
\end{itemize}

\subsubsection{YOLOv9~\cite{re_add_8_YOLOv9}}
YOLOv9 introduces two novel techniques, including programmable gradient information (PGI) and generalized efficient layer aggregation network (GELAN). The descriptions of PGI and GELAN are as follows:
\begin{itemize}
 \item The PGI theory involves redesigning a technique called the auxiliary reversible branch that allows the YOLOv9 to generate reliable gradient information during the training phase and pass it to the main branch. 

\item GELAN combines the functions of the CSPNet and ELAN networks. 
It splits the input of each computational unit into two parts: one part is directly passed to the output, while the other part is transformed by the computational unit. 
Then, these two parts of computation are combined to form the final output. 
Thus, GELAN integrates information between computational units while enhancing flexibility and generalization. 
\end{itemize}

The proposed model consists of two parts: data augmentation with ConSinGAN and training the YOLO model. The limitations of the YOLO model are as follows:
\begin{itemize}
    \item Computational capability: Depending on the constraints of hardware, it is necessary to use a lightweight YOLO model, which may sacrifice accuracy.
    \item Training cost: To achieve accurate detection, the YOLO model requires a large amount of labeled data for training because collecting and labeling sufficient training data is a time-consuming and costly task.
    \item External factors: In the image, additional obstacles or illumination changes may reduce the accuracy of the YOLO model.
    \item Size of defect features: If the defect features are tiny, the single-stage detection is inferior to multi-stage detection.
\end{itemize}

\subsection{Performance Evaluation}
\label{subsec:Performance evaluation}
The leave-one-out cross-validation (LOOCV) is used to measure the performance of the proposed DIP detection system. The testing set includes one subject’s data, while the training set uses the remaining subjects’ data. LOOCV is repeated $k$ times, where $k$ is the total number of subjects, until all subjects have been used as the testing set. By using $k$-folds, the average performance can be obtained.
 
In our experiments, we adopt a confusion matrix as performance measurement, and five evaluation metrics, precision ($PRE$), recall ($REC$),  F1-Score ($F1$), false positive rate ($FPR$), and true negative rate ($TNR$) as shown from Eq.~\eqref{eq7} to \eqref{eq11} to measure the detection performance.
\begin{equation}
        PRE= \rm\frac{TP}{TP+FP},
        \label{eq7}
\end{equation}
\begin{equation}
        REC= \rm\frac{TP}{TP+FN},
        \label{eq8}
\end{equation}

\begin{equation}
        F1= 2\times\frac{PRE \cdot REC}{PRE+REC},
        \label{eq9}
\end{equation}
\begin{equation}
        FPR= \rm\frac{FP}{FP+TN},
        \label{eq10}
\end{equation}
\begin{equation}
        TNR= \rm\frac{TN}{FP+TN},
        \label{eq11}
\end{equation}
where TP (true positive), TN (true negative), FP (false positive), and FN (false negative) represent the cases: the defect and normal DIPs labeled as such are correctly recognized, the normal DIPs labeled as defective DIPs are misclassified, and the defective DIPs labeled as normal DIPs are misclassified, respectively.

The performance evaluation of detection depends on the intersection over the union (IoU), and mean average precision (mAP), where mAP0.5 means the threshold of YOLO is set to 0.5.
All model training is conducted on a PC with an Intel 13th Gen I7-13700K CPU (3.40 GHz), NVidia RTX4080 graphics card with 16 GB of dedicated memory, running on a 64-bit Windows 10 operating system.
The training and testing phase for the models is executed in an Anaconda (a Python derivative) environment, and the SCADA interface is integrated using Visual Studio 2019.

\section{Experimental Results}
\label{sec:Experimental results}

In our experiments, the nine kinds of models are specified as follows: 1) Threshold-based detection, 2) YOLOv3, 3) YOLOv4, 4) YOLOv7, 5) YOLOv9, 6) YOLOv3 with ConSinGAN, 7) YOLOv4 with ConSinGAN,  8) YOLOv7 with ConSinGAN, 9) YOLOv9 with ConSinGAN. The hyperparameter settings of these models are as follows:

\begin{itemize}
    \item The threshold-based detection: Based on Eq.~\eqref{eq4}, $TB$ is assigned a constant value of 100.
    \item The proposed YOLOv3, v4, v7, and v9 models with/without ConSinGAN: Batches = 32, images = 416$\times$416, learning rate = 0.001, and maximum batches = 10000.
\end{itemize}
    The hyperparameter settings for the generative adversarial network ConSinGAN are as follows: Learning rate = 0.1, the number of trained stages = 10, and the original images are captured using the industrial camera described in Sec.~\ref{sec:Automated system development}. The total number of original images is 672. The ConSinGAN selects images with significant defective features for image augmentation. The total number of generated images is 3,183.
    
\subsection{DIP images dataset}
\subsubsection{ConSinGAN Augmentation}
\label{subsec:Image augmentation}
\begin{figure*}[ht]
    \centerline{\includegraphics[width=1.2\columnwidth]{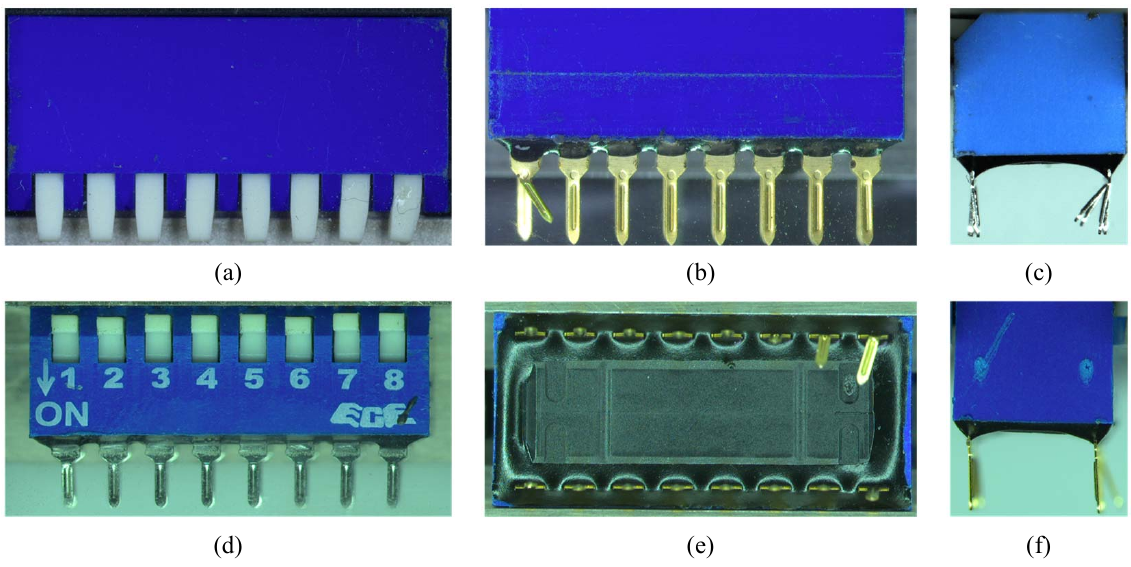}}
    \caption{Original images of defective DIP, (a) top, (b) back, (c) right, (d) front, (e) bottom, and (f) left sides.}
    \label{fig11:Original defect images}
\end{figure*}

\begin{figure*}[ht]
    \centerline{\includegraphics[width=1.2\columnwidth]{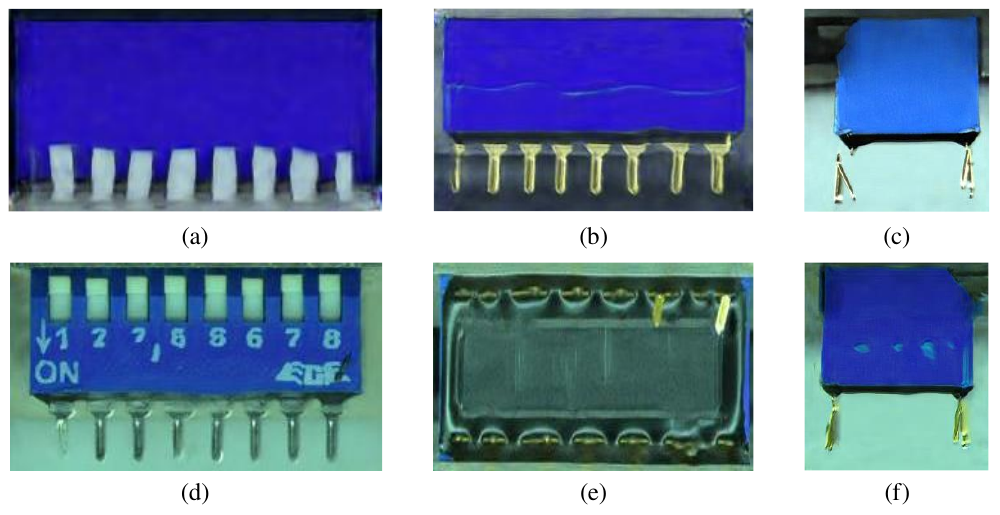}}
    \caption{Images generated by ConSinGAN, (a) top, (b) back, (c) right, (d) front, (e) bottom, and (f) left sides.}
    \label{fig12:Images generated by ConSinGAN}
\end{figure*}

Initially, GAN and DCGAN were used to augment data.
However, the training performances of GAN and DCGAN were poor because of the limited number of defective image samples.
Therefore, ConSinGAN is used, which generates similar images from a smaller number of images for data augmentation.
The original images are shown in Fig.~\ref{fig11:Original defect images}, and the augmented results are shown in Fig.~\ref{fig12:Images generated by ConSinGAN}.
The amount of each side of the defective DIP in the generated dataset is shown in Table~\ref{tb3:Generated image data statistics}.

\begin{table}[h]
\centering
\caption{Statistics of ConSinGAN-generated images.}
\scalebox{0.8}{
      \begin{tabular}{|c|c|c|c|c|c|c|c|}
      \hline
       \textbf{\makecell{Type of\\ defect}} & \textbf{Front} & \textbf{Back} &\textbf{Top} & \textbf{bottom} & \textbf{Left} & \textbf{Right} & \textbf{Total}\\\hline
       \textbf{\makecell{Surface \\ defect}} & 340 & 228 &209 & 183 & 310 & 346 & 1616\\\hline
       \textbf{\makecell{Pin \\ defect}} & 321 & 216 &201 & 177 & 325 & 327 & 1567\\\hline
    \end{tabular}}
\label{tb3:Generated image data statistics}
\end{table}

\subsection{Performances of YOLO with/without ConSinGAN}
\label{subsec:YOLO Deep learning}

The YOLO model is trained for each aspect of the generated image data. These 3,183 images are divided into two categories: surface defects (surface overflow, surface scratches, and surface dirt) and pin defects (pin misalignment). We compare YOLOv3, v4, v7, and v9 with/without the ConSinGAN augmented sets to identify the optimal model. The training set is conducted on 80\% of the dataset, i.e., 2,547 images. The validation set is adopted on the other 10\% (318 images). The remaining 10\% (318 images) are used for the testing set. Within this dataset, we define two distinct defect types that include: 1) Surface defects account for approximately 50.7\% of defects, and 2) Misaligned pins account for 49.3\%. Table~\ref{tb:Yolo model results} demonstrates the performances of YOLOv3, v4, v7, and v9 with/without the ConSinGAN in terms of mAP rate,  $PRE$, $REC$, $F1$, $FPR$, and $TNR$ evaluations of all six YOLO models.

\begin{table*}[h]
\centering
\caption{Performance Comparison of YOLO models.}
\scalebox{1}{
      \begin{tabular}{|c|c|c|c|c|c|c|c|}
      \hline
       \textbf{Model} & mAP0.5 & \textbf{$PRE$} &\textbf{$REC$} &\textbf{$F1$} &\textbf{$FPR$} &\textbf{$TNR$} &\textbf{Detection time (ms)}\\\hline
       \textbf{YOLOv3} & 66.2\% & 73.2\% &51.5\% &61.8\% &35.0\% &64.9\% &322\\\hline
       \textbf{YOLOv4} & 65.4\% & 80.1\% &61.4\% &68.9\% &35.2\% &64.8\% &314\\\hline
       \textbf{YOLOv7} & 65.1\% & 84.2\% &58.6\% &68.1\% &34.3\% &65.6\% &297\\\hline
        \textbf{YOLOv9} & 74.5\% & 79.2\% &66.4\% &72.8\% &40.3\% &59.7\% &279\\\hline
       \textbf{YOLOv3 with ConSinGAN} &91.9\% & 92.8\% &91.4\% &92.4\% &8.9\% &91.1\% & 321\\\hline
       \textbf{YOLOv4 with ConSinGAN} & 89.1\% & 91.3\% &92.6\% &91.5\% &10.7\% &89.2\% &310\\\hline
       \textbf{YOLOv7 with ConSinGAN} & \textbf{95.5\%} & \textbf{94.9\%} &\textbf{90.7\%} &\textbf{93.4\%} &\textbf{3.7\%} &\textbf{96.4\%} &285\\\hline
       \textbf{YOLOv9 with ConSinGAN} & 90.5\% & 87.7\% &82.9\% &85.0\% &9.8\% &90.2\% &\textbf{278}\\\hline       
    \end{tabular}}
  
\label{tb:Yolo model results}
\end{table*}

\begin{figure}[htbp]
    \centerline{\includegraphics[width=1\columnwidth]{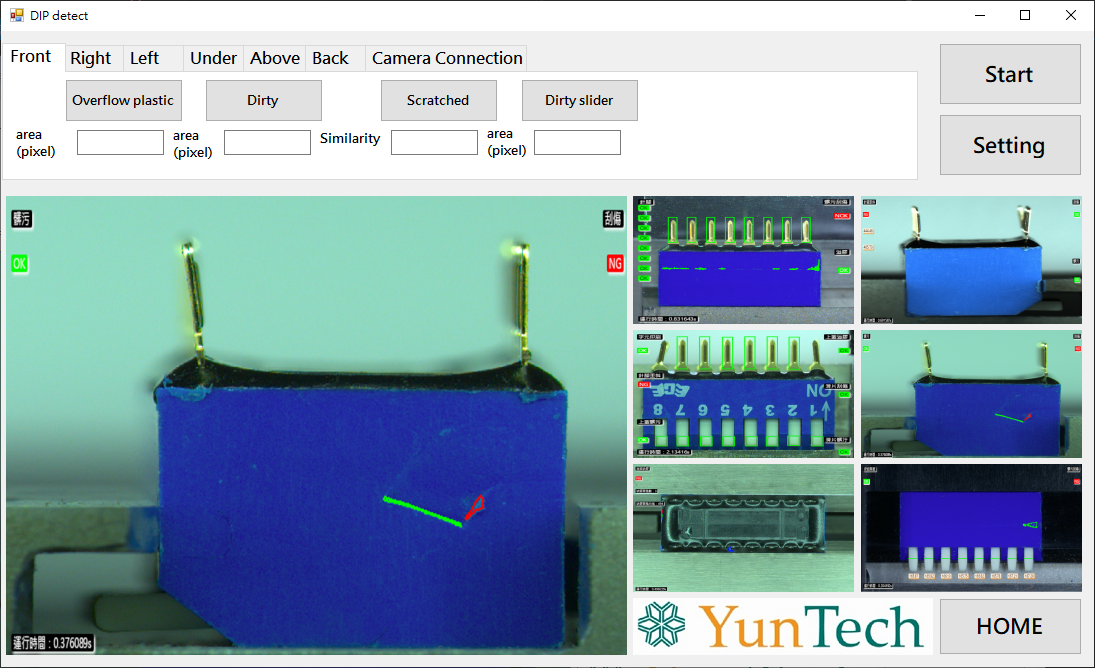}}
    \caption{The SCADA interface.}
\label{fig14:UI}
\end{figure}

As shown in Table~\ref{tb:Yolo model results}, 1) With ConSinGAN, YOLOv3, v4, v7, and v9 outperform the models without ConSinGAN in terms of accuracy metrics due to data augmentation. By data augmentation, the YOLOv3, v4, v7, and v9 have been enhancing the performance in the training phase because of the increase in the amount of training data. 2) YOLOv7 with ConSinGAN achieves outstanding performance on accuracy metrics and outperforms the current newest YOLOv9 with ConSinGAN. The proposed YOLOv7 with ConSinGAN model achieves an accuracy in terms of mAP0.5 = 95.5\% and outperforms the other models.
However, the detection speed of YOLOv9 with ConSinGAN outperforms the other models and is slightly better than YOLOv7 with ConSinGAN. The detection time of each model is approximately 300 ms. Therefore, we select the YOLOv7 model in a DIP image detection system on the SCADA interface for the user, as illustrated in Fig.~\ref{fig14:UI}. In addition, ConSinGAN is only used to augment the dataset. Thus, it does not impact the complexity of YOLO models.

\subsection{Performances of Threshold-based Detection}
\label{subsec:Automation system}
For baseline comparison, we introduce the performance evaluation of threshold-based detection.
Table~\ref{tb5:Threshold based method} presents the number of images, accuracy, and detection time from different sides of DIP by threshold-based detection.
Each workpiece is inspected from all six sides. 
The number of images is the same as in the YOLO experiments.
The threshold value is set to 0.5.
The accuracy is calculated by
\begin{equation}
        A = \frac{DT - F}{DT}\times 100\%,
        \label{eq12}
\end{equation}
where $A$ represents the accuracy, $DT$ is the
total number of detections, $F$ is the number of detection errors.

Table~\ref{tb5:Threshold based method} shows the average accuracy is $87.81\%$ on six-sided DIP image detection. The total detection time is 3.807 seconds, which matches the detection time. Our experiment involves an actual production line machine, divided into six-sided detectors that are moved synchronously. Therefore, the detection time is determined by the station with the longest detection time. The proposed YOLOv7 with ConSinGAN model outperforms the threshold-based system, achieving an accuracy of 95.5\%.

\subsection{Comparison of Detection Speed}

\begin{figure}[h]
    \centerline{\includegraphics[width=1\columnwidth]{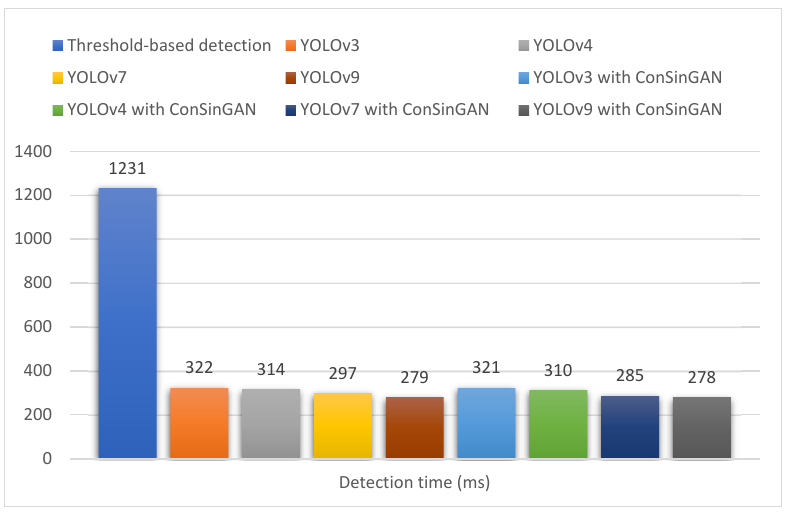}}
    \caption{Comparison of detection speed.}
\label{fig15:Comparison}
\end{figure}

Fig.~\ref{fig15:Comparison} demonstrates the detection time of the YOLO models and shows a significantly higher detection speed than the threshold-based system. The threshold-based method needs to reset the parameters for every six-sided DIP image. However, YOLO models only need one setting, because the training and testing stages are both performed using six-sided DIP images as input. Different from threshold-based image processing, YOLO models do not require parameter adjustments to different sides of DIP to obtain accurate results. 
Thus, there are performance gaps between YOLO models and the threshold-based method. From Fig.~\ref{fig15:Comparison}, it is feasible to replace the threshold-based image-processing defect detection rate with the proposed YOLOv7 with the ConSinGAN approach.
\begin{table}[h]
\centering
\caption{Performance of the threshold-based detection}
\scalebox{1}{
      \begin{tabular}{|c|c|c|c|}
      \hline
       \textbf{Detected side} & \textbf{Test number} & \textbf{Accuracy} &\textbf{Detection time (ms)} \\\hline
       \textbf{Left} & 635 & 86.30\% &290 \\\hline
       \textbf{Top} & 410 & 86.58\% &391 \\\hline
       \textbf{Right} & 673 & 88.85\%&359 \\\hline
       \textbf{bottom} & 360 & 92.22\% &815 \\\hline
       \textbf{Back} & 444 & 90.54\% &721 \\\hline
       \textbf{Front} & 661 & 84.56\% &1231 \\\hline
       \textbf{Total} & 3183 & 87.81\% &3807 \\\hline
    \end{tabular}}
\label{tb5:Threshold based method}
\end{table}

\section{Conclusion}
\label{sec:Conclusion}
In this paper, our goal is to establish an AOI system for defective DIP detection to improve the quality of production and save human efforts.
We use the SCADA interface to integrate image detection and automated mechanism hardware for a large number of DIP inspections.
For the image detection model, we use threshold-based detection and DL-based YOLO models.
The threshold-based detection outperforms the YOLO models in terms of accuracy.
However, the threshold-based detection is time-consuming.
Thus, we adopt ConSinGAN to augment the dataset for YOLO training and improve the accuracy.
With ConSinGAN, YOLOv7 achieves an accuracy of $95.5\%$, outperforming the accuracy of $87.81\%$ by threshold-based detection.
The detection times of different YOLO models are 285 to 322 ms, which outperforms 1231 ms of the threshold-based detection. 
In summary, ConSinGAN effectively enhances the accuracy of YOLO models while maintaining the outstanding detection time. 
For future works, we summarize as follows:
\begin{itemize}
    \item Applying future novel models to improve the detection model network.
    \item Increasing the variety of workpieces for the automated detection system, such as rotating DIP and surface mount devices, which are mass-produced electronic components.
    \item Expanding the SCADA system to cover additional production lines, aiming for smart factories and AIoT integration.
\end{itemize}

\begin{IEEEbiography}[{\includegraphics[width=1.3in,height=1.7in,clip,keepaspectratio]{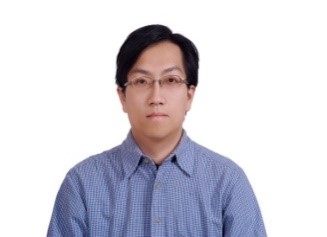}}]{Wei–Lung Mao}was born in Taiwan, in 1972. He received his B.S. in electrical engineering from the National Taiwan University of Science and Technology (NTUST), Taipei, Taiwan, in 1994. He earned his M.S. and Ph.D. degrees in Electrical Engineering at the National Taiwan University (NTU), Taipei, Taiwan, in 1996 and 2004. 

He is currently a Professor in the Department of Electrical Engineering and Graduate School of Engineering Science and Technology, National Yunlin University of Science and Technology (NYUST), Yunlin, Taiwan. His research interests are satellite navigation systems, intelligent and adaptive control systems, adaptive signal processing, neural networks, and precision control. 
\end{IEEEbiography}

\begin{IEEEbiography}[{\includegraphics[width=1in,height=1.25in,clip,keepaspectratio]{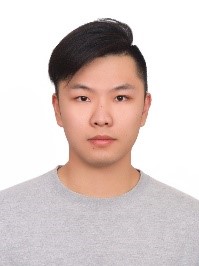}}]{Chun-Chi Wang}was born in Taiwan in 1995 and received his B.A. from the Department of Materials and Energy Engineering, MingDao University, Changhua, Taiwan, in 2018. He earned his M.S. at the Graduate Institute of Aeronautical and Electronic Technology from National Formosa University (NFU), Yunlin, Taiwan, in 2021. He is currently a PhD student at the National Yunlin University of Science and Technology (NYUST), Yunlin, Taiwan. 

His research interests include automated image inspection, deep learning, robotic arm control systems, intelligent manufacturing production lines, and EtherCAT.
\end{IEEEbiography}

\begin{IEEEbiography}[{\includegraphics[width=1in,height=1.25in,clip,keepaspectratio]{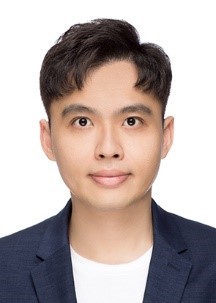}}]{Po-Heng Chou}(Member, IEEE) was born in Tainan, Taiwan. He received the B.S. degree in electronic engineering from National Formosa University (NFU), Huwei, Yunlin, Taiwan, in 2009, the M.S. degree in communications engineering from National Sun Yatsen University (NSYSU), Kaohsiung, Taiwan, in 2011, and the Ph.D. degree from the Graduate Institute of Communication Engineering (GICE), National Taiwan University (NTU), Taipei, Taiwan, in 2020. 
His research interests include AI for communications, deep learning-based signal processing, wireless networks, and wireless communications, etc.

He was a Postdoctoral Fellow at the Research Center for Information Technology Innovation (CITI), Academia Sinica, Taipei, Taiwan, from Sept. 2020 to Sept. 2024. 
He was a Postdoctoral Fellow at the Department of Electronics and Electrical Engineering, National Yang Ming Chiao Tung University (NYCU), Hsinchu, Taiwan, from Oct. to Dec. 2024.
He has been elected as the Distinguished Postdoctoral Scholar of CITI by Academia Sinica from Jan. 2022 to Dec. 2023. He is invited to visit Virginia Tech (VT) Research Center (D.C. area), Arlington, VA, USA, as a Visiting Fellow, from Aug. 2023 to Feb. 2024.
He received the Partnership Program for the Connection to the Top Labs in the World (Dragon Gate Program) from the National Science and Technology Council (NSTC) of Taiwan to perform advanced research at VT Institute for Advanced Computing (D.C. area), Alexandria, VA, USA, from Jan. 2025 to present.

Additionally, Dr. Chou received the Outstanding University Youth Award and the Phi Tau Phi Honorary Membership from NTU in 2019 to honor his impressive academic achievement. He received the Ph.D. Scholarships from the Chung Hwa Rotary Educational Foundation from 2019 to 2020.
\end{IEEEbiography}

\begin{IEEEbiography}[{\includegraphics[width=1in,height=1.25in,clip,keepaspectratio]{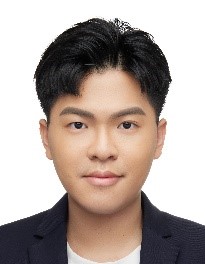}}]{Yen-Ting Liu} was born in Kaohsiung, Taiwan, in 2000. He received the B.S. degree in computer and communications engineering from the National Kaohsiung University of Science and Technology (NKUST), Kaohsiung, Taiwan, in 2022. 

He is currently studying for the M.S. degree in communications engineering from National Sun Yat-sen University (NSYSU), Kaohsiung, Taiwan. His research interests include deep learning in wireless communications, communication theory, and resource allocation for wireless networks.
\end{IEEEbiography}

% that's all folks
\end{document}